\documentclass{article}
\usepackage[left=3cm, right=3cm, top=2cm]{geometry}
\usepackage{graphicx}
\usepackage{array}
\usepackage{nicefrac}
\usepackage{amsfonts}
\usepackage{mathtools}
\usepackage{hhline}
\usepackage{amsmath,amssymb}
\usepackage{amsbsy}
\usepackage{accents}

\usepackage[table,xcdraw]{xcolor}
\usepackage{lineno}
\usepackage{lipsum}                     
\usepackage{xargs}

\usepackage{subcaption}
\usepackage{color}
\usepackage{multirow}
\usepackage{soul}
\soulregister\cite7
\soulregister\citep7
\soulregister\citet7
\soulregister\ref7
\soulregister\pageref7
\usepackage{ulem}

\newcommand{\tabincell}[2]{\begin{tabular}{@{}#1@{}}#2\end{tabular}}

\title{Content-Aware Inter-Scale Cost Aggregation \\ for Stereo Matching}

\author{Chengtang Yao, Yunde Jia, Huijun Di, Yuwei Wu, Lidong Yu}
\date{
\small
School of Computer Science, Beijing Institute of Technology, Beijing.
}

\begin{document}

\maketitle


\begin{abstract}
Cost aggregation is a key component of stereo matching for high-quality depth estimation. Most methods use multi-scale processing to downsample cost volume for proper context information, but will cause loss of details when upsampling. In this paper, we present a content-aware inter-scale cost aggregation method that adaptively aggregates and upsamples the cost volume from coarse-scale to fine-scale by learning dynamic filter weights according to the content of the left and right views on the two scales. Our method achieves reliable detail recovery when upsampling through the aggregation of information across different scales. Furthermore, a novel decomposition strategy is proposed to efficiently construct the 3D filter weights and aggregate the 3D cost volume, which greatly reduces the computation cost. We first learn the 2D similarities via the feature maps on the two scales, and then build the 3D filter weights based on the 2D similarities from the left and right views. After that, we split the aggregation in a full 3D spatial-disparity space into the aggregation in 1D disparity space and 2D spatial space. Experiment results on Scene Flow dataset, KITTI2015 and Middlebury demonstrate the effectiveness of our method.
\end{abstract}

\section{Introduction}
\begin{figure}[tp]
	\centering
	\includegraphics[width=.95\textwidth]{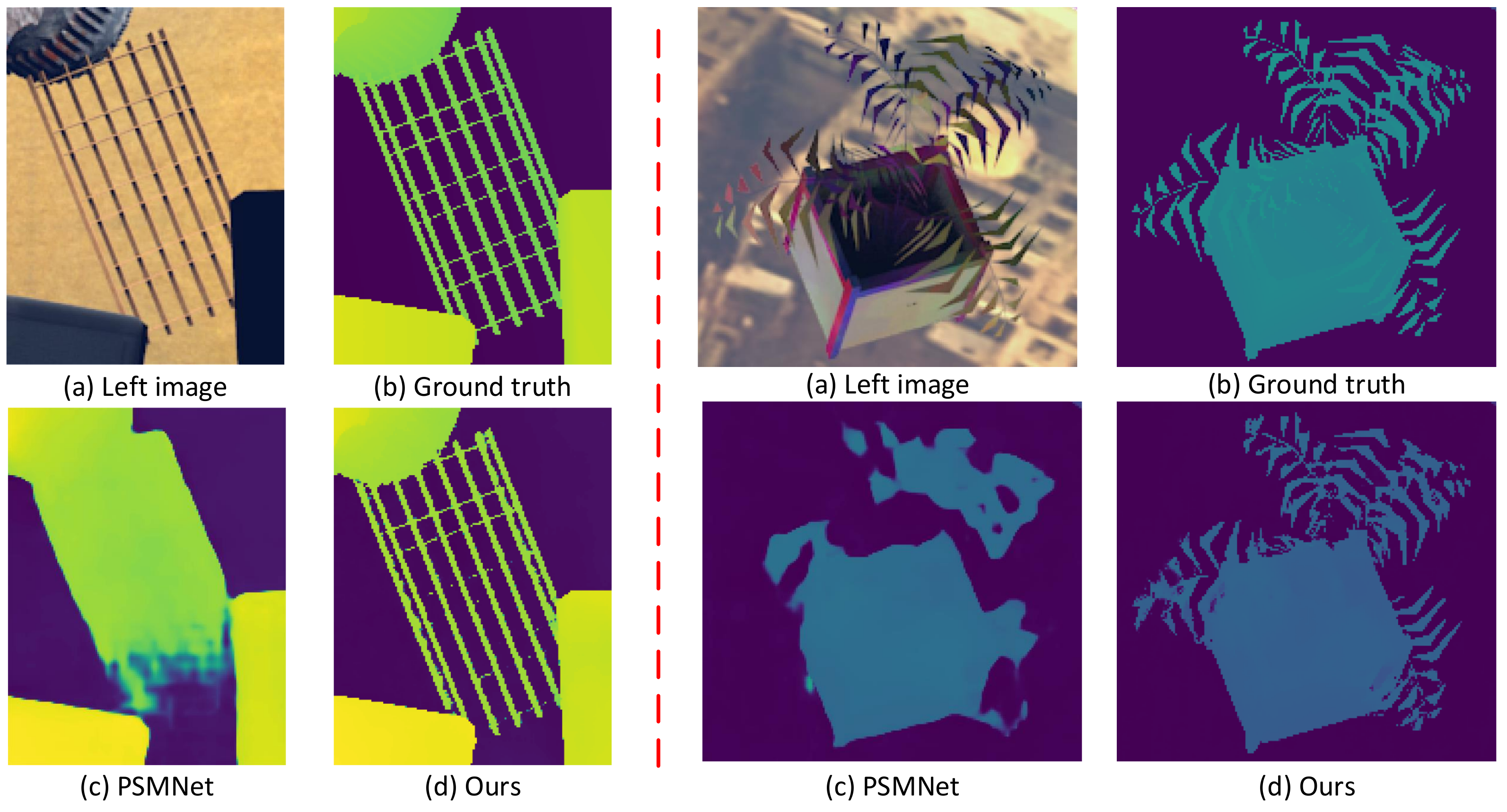}
	\centering
    \caption{Performance illustrations. The PSMNet \cite{chang2018pyramid} uses deconvolution and bilinear interpolation with fixed mapping weights for cost volume upsampling. Our method learns content-aware filter weights for inter-scale cost aggregation, and achieves much finer details.}
    \label{Fig:abstract}
\end{figure}
Cost aggregation is a key component of stereo matching \cite{scharstein2002taxonomy}, which filters cost volume to rectify the mismatched pixels via the context information within the support window of each pixel. Most existing aggregation methods \cite{marr1979computational,jen2011adaptive,hu2013comparisons,kendall2017end,chang2018pyramid,guo2019group,duggal2019deeppruner,wu2019semantic,zhang2014cross} usually incorporate multi-scale processing to adjust the receptive field of filters to provide appropriate context information for areas with different sizes. Under multi-scale processing, a cost volume filtered at coarse scale is needed to be upsampled to fine scale. Those methods simply upsample a cost volume through nearest-neighbor/bilinear interpolation or deconvolution that uses fixed mapping weights over the entire volume. However, it is difficult to use fixed weights of mapping to adaptively recover the details that are poorly represented at coarse scales.

In this paper, we present a novel method of content-aware inter-scale cost aggregation that adaptively aggregates and upsamples the cost volume from coarse-scale to fine-scale by learning dynamic filter weights according to the content of the left and right views on the two scales. Our method achieves reliable detail recovery when upsampling through the aggregation of information across different scales. In our method, content-aware weights of filter for each site in a fine-scale cost volume are learned to adaptively aggregate information and to amplify remaining details in a coarse-scale cost volume, see Figure \ref{Fig:abstract} for the results achieved by our method.

\begin{figure}[htbp]
	\centering
	\includegraphics[width=.9\textwidth]{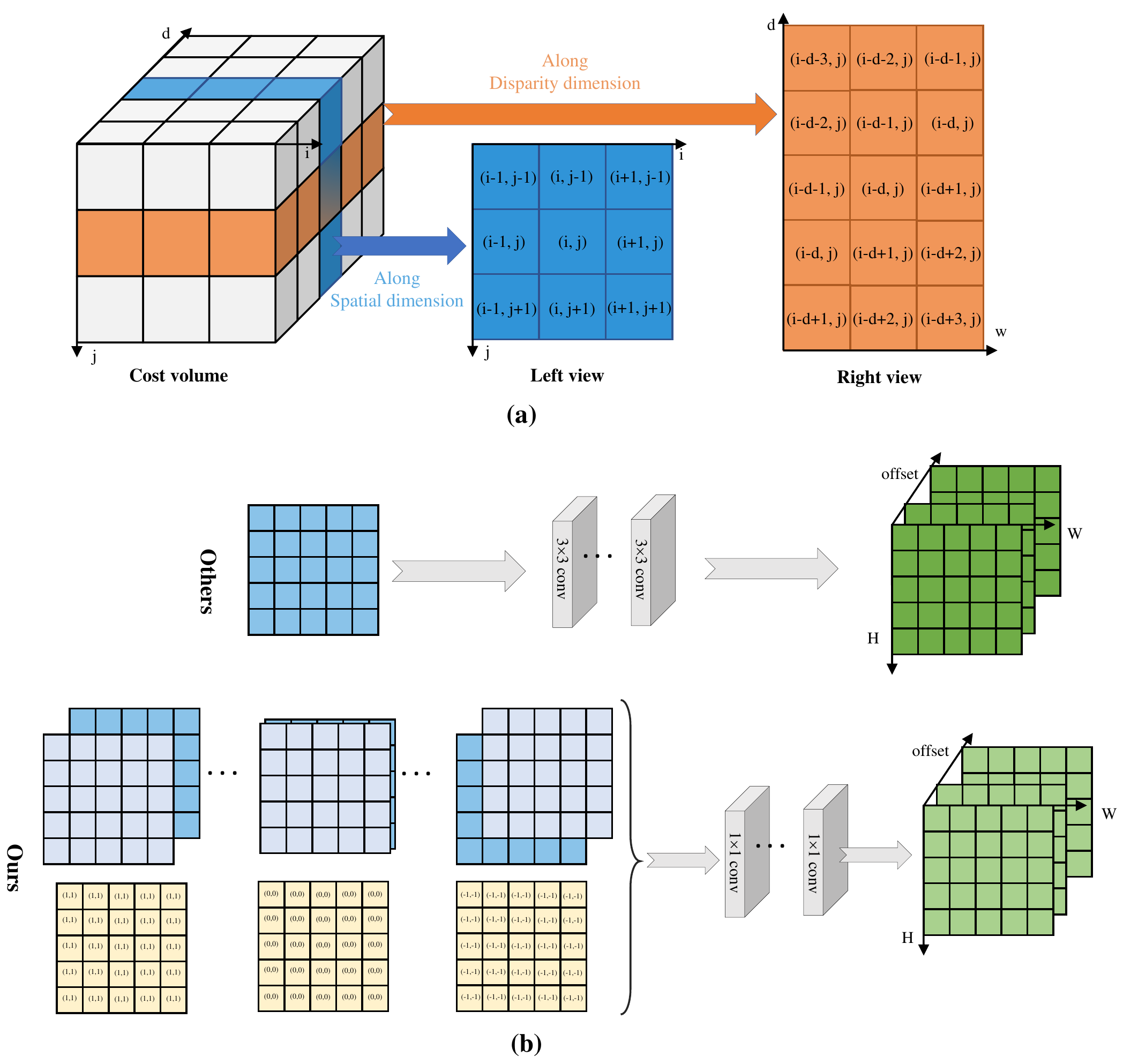}
	\centering
    \caption{Illustration of our strategy of effective content-aware weight learning. (a) Relationship between 3D cost volume and stereo views. (b) Our paired feature inputs with proper spatial relationship encoding (the second row) vs. unitary feature input in existing work for pairwise weight learning (the first row).}
    
    \label{Fig:differences}
\end{figure}

Our method uses a novel strategy of effective content-aware weight learning that contains the following ideas: (\romannumeral1) We use left and right views, which we call as stereo views for convenience, to learn content-aware filter weights for cost aggregation. In contrast, existing content-aware filters for cost aggregation use only single-view information (i.e. features from the left view) to learn the weights of guided filters \cite{zhang2019ga} or affinity kernels \cite{cheng2019learning}. A cost volume is about 3D information from stereo views, as shown in Figure \ref{Fig:differences}(a). Compared to the guidance based on 2D textural information from a single view, cost aggregation according to the guidance based on 3D information from stereo views can achieve better results. However, it is not straightforward to exploit stereo views for weight learning. A direct concatenation of feature maps from stereo views for weight learning cannot work well, since their unaligned contents will undermine each other's guidance. We will explain later how we use stereo views. (\romannumeral2) Each content-aware weight of a filter represents the similarity between two points with specific spatial relationship. Existing methods use a series convolutions with large receptive field on unitary feature input to learn the content-aware weights \cite{jia2016dynamic,jo2018deep,cheng2019learning,wang2019carafe,liu2017learning,zhang2019ga,mazzini2018guided}, which we call as implicit spatial relationship encoding, as shown in Figure \ref{Fig:differences}(b). Differently, we explicitly encode the spatial relationship to learn the content-aware weights, including the shifting of feature maps from two scales and the construction of location map, which we call as explicit spatial relationship encoding. In more detail, the feature maps from two scales are shifted relatively according to each spatial relationship and then concatenated as the input for learning of specific weights. Location maps about the relative position information are also introduced into the input. As each content-aware weight is related to the specific spatial relationship of two pixels, it is better to learn this pairwise relationship from paired feature inputs, rather than only from unitary feature input. What's more, it becomes possible to use only $1 \times 1$ convolutions to generate the content-aware weights, which is more efficient than previous methods. The effectiveness of our strategy is shown in the ablation study.

Our method also uses a decomposition strategy to efficiently construct the 3D filter weights and aggregate the 3D cost volume. The mechanism avoids content-aware weight learning and aggregation in a full 3D spatial-disparity space that will cause huge computation cost. Instead of directly learning 3D filter weights for the aggregation on 3D cost volume, we only learn 2D similarities from adjacent scales for the left and right views. 3D content-aware weights are constructed on-the-spot from those 2D similarities via warping with corresponding disparity. Both memory and computation can be reduced significantly in such a way. To further reduce the computation cost, we split the inter-scale aggregation in a full 3D spatial-disparity space into the aggregation in 1D disparity space and the aggregation in 2D spatial space. As a result, our method could efficiently construct and use 3D content-aware weights for inter-scale cost aggregation.

Our contributions are summarized as follows:
\begin{itemize}
\item We present a novel method of content-aware inter-scale cost aggregation that can not only upsample the cost volume for reliable detail recovery, but also aggregate the information across scales to improve the quality of disparity estimation.

\item We present a novel strategy of effective content-aware weight learning for inter-scale aggregation on 3D cost volume, by using 3D information from stereo views and by explicit spatial relationship encoding between two scales.

\item We present a novel mechanism of 3D filter weight construction and 3D cost aggregation decomposition, to efficiently construct and use 3D content-aware weights for inter-scale cost aggregation. Both memory consumption and computation cost can be reduced significantly.
\end{itemize}

\section{Related Work}
\subsection{Multi-Scale Cost Aggregation}
Most existing methods of multi-scale cost aggregation improve the stereo matching results via local context information with coarse-to-fine strategy \cite{marr1979computational,hu2013comparisons,jen2011adaptive}. 
Zhang et al. \cite{zhang2014cross} argue that the information from different scales is complementary to each other, and they present cross-scale cost aggregation that fuses the cost volume from all scales simultaneously to achieve inter-scale consistency. Recently, it is popular to embed cost aggregation into deep stereo network for ene-to-end learning. Kendall et al. \cite{kendall2017end} first propose to incorporate cost aggregation into end-to-end deep stereo network via multi-scale 3D convolution. Chang and chen \cite{chang2018pyramid} use three stacked hourglass network to enlarge the receptive field for multi-scale cost aggregation. Yu et al. \cite{yu2018deep} propose an explicit cost aggregation to select the best cost proposal. In order to relieve the influence of challenging regions, Zhang et al. \cite{zhang2019ga} formulate traditional SGM into deep network and propose a semi-global aggregation layer and a local guided aggregation layer. For better details and lower noises, Cheng et al. \cite{cheng2019learning} propose to propagate the information within the disparity space and scale space. The above methods use fixed kernels/weights to map the cost volume to next scale. However, the fixed kernels/weights would cause serious loss of details, as it is difficult to use fixed mapping to selectively enhance the details poorly represented at coarser scales. In contrast, our work focuses on content awareness, which can naturally adapt the mapping to different areas and dynamically recover the details.

\subsection{Guided Upsampling}
As one of the most important components in deep neural network, guided upsampling is popular in many dense tasks, such as super resolution \cite{jo2018deep}, segmentation \cite{he2019adaptive} and depth upsampling \cite{hui2016depth}. Most of those tasks use bilinear interpolation to upsample the input and then filter the result under guidance. This two-stage processing needs additional computation cost compared to one-stage processing. The receptive field of aggregation is also limited at finer-scale and is smaller than directly sampling from coarser-scale.

In one-stage guided upsampling, Shi et al. \cite{shi2016real} learn convolution filters to replace the handcrafted bicubic filter. Tian et al. \cite{tian2019decoders} design a data-dependent upsampling to replace the bilinear operation, which learns an inverse linear projection via reconstruction error. Mazzini et al. \cite{mazzini2018guided} point out that it is easy to predict wrong labels on object boundaries using vanilla upsampling. Thus they predict a high-resolution pixel-specific guidance offset table for upsampling. Wang et al. \cite{wang2019carafe} argue that nearest neighbor and bilinear interpolation cannot capture the rich semantic information and propose to generate the mask from input to achieve instance-specific awareness.

In our work, we address the problem of fixed weights leading to the loss of details in cost volume mapping, and then build adaptive kernels to fit the mapping on different areas. Our work is different from aforementioned methods as following. (\romannumeral1) We explicitly encode spatial relationship to extract the mapping between fine-scale and coarse-scale feature maps as guidance, while previous guidance generation methods only depend on the filtering effects over the feature maps from single scales or the direct concatenation of feature maps. The superiority of explicit spatial encoding scheme is proved in our experiments. (\romannumeral2) We further consider the stereo views in stereo matching and leverage the guidance from two views to achieve the content awareness, while prior methods only have one view. The effectiveness of stereo views is also verified in our ablation study. (\romannumeral3) Previous guided upsampling methods are based on 2D feature maps. It would cost huge computation resources to directly apply them to 3D cost volume. Differently, our method is much more efficient, as we build the 3D kernels on-the-spot from 2D guidance and split 3D spatial-disparity space to 1D disparity space and 2D spatial space separately.

\subsection{Content-Aware Filtering}
Content-aware filters are usually built from a series of convolutions on the reference feature maps. Jia et al. \cite{jia2016dynamic} propose to learn the filter conditioned on inputs. Zhang et al. \cite{zhang2019ga} and Cheng et al. \cite{cheng2019learning} learn the affinity along each direction via filtering effects as shown in Figure \ref{Fig:differences}(b). Other methods, like super resolution \cite{jo2018deep}, segmentation \cite{he2019adaptive} and depth upsampling \cite{hui2016depth}, also depend on the filtering effects to learn the weights along specific direction. In contrast, we explicitly encode pairwise spatial relationships into the input as presented in Figure \ref{Fig:differences} (b). The explicit scheme has lower computation cost than the above methods as we use shared convolutions with size of $1 \times 1$, while the above methods need different series of convolutions with size of at least $3 \times 3$. Also, instead of using the single left view, we learn the content-aware filter weights for cost aggregation via stereo views which are inherent in cost volume as shown in Figure \ref{Fig:differences} (a).

\begin{figure}[htbp]
	\centering
	\begin{subfigure}{.9\textwidth}
	    \centering
        \begin{minipage}[t]{0.98\linewidth}
        \centering
        \includegraphics[width=.95\textwidth]{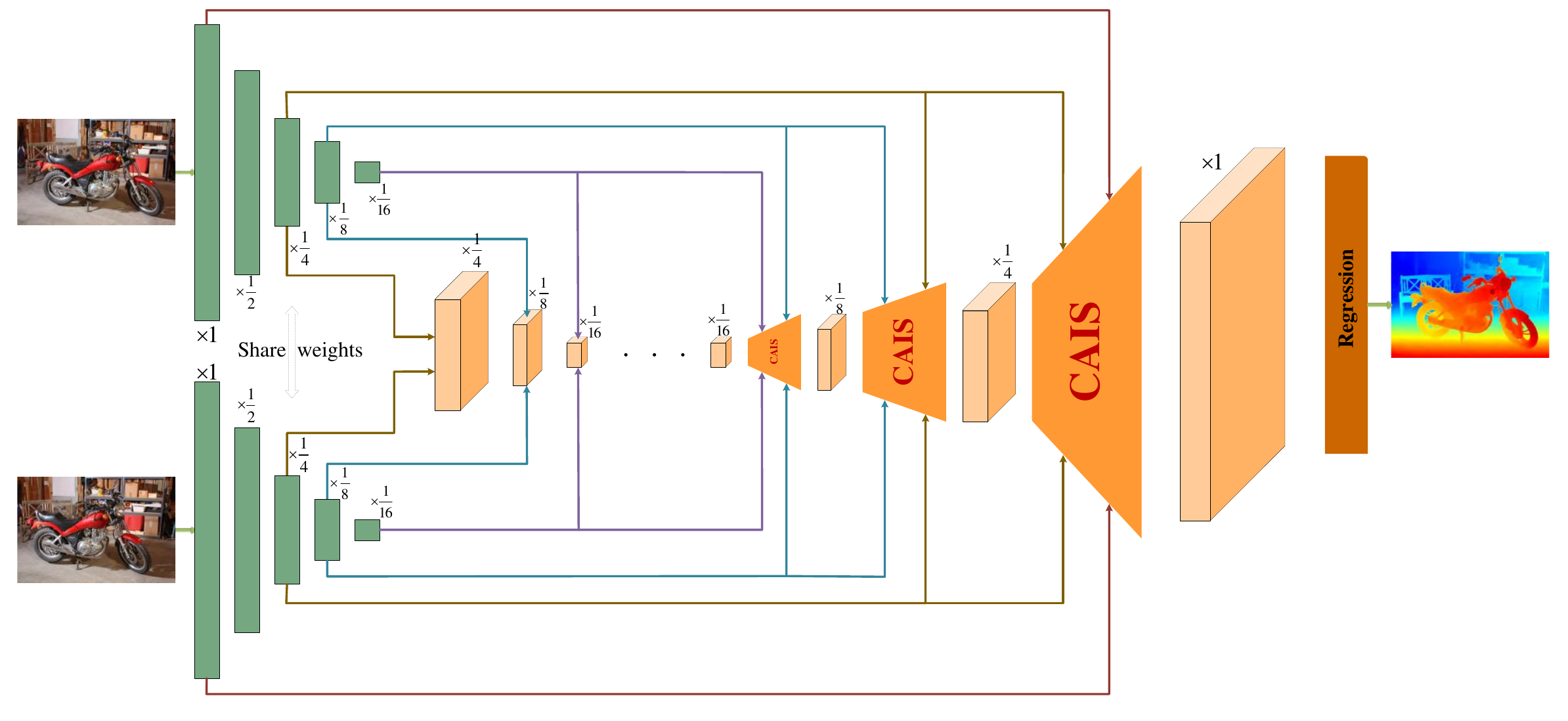}
        \caption{The visualization of embedding our method into the current multi-scale framework of stereo network.}
        \end{minipage}%
    \end{subfigure}
    
	\begin{subfigure}{.9\textwidth}
	    \centering
        \begin{minipage}[t]{0.98\linewidth}
        \centering
        \includegraphics[width=.95\textwidth]{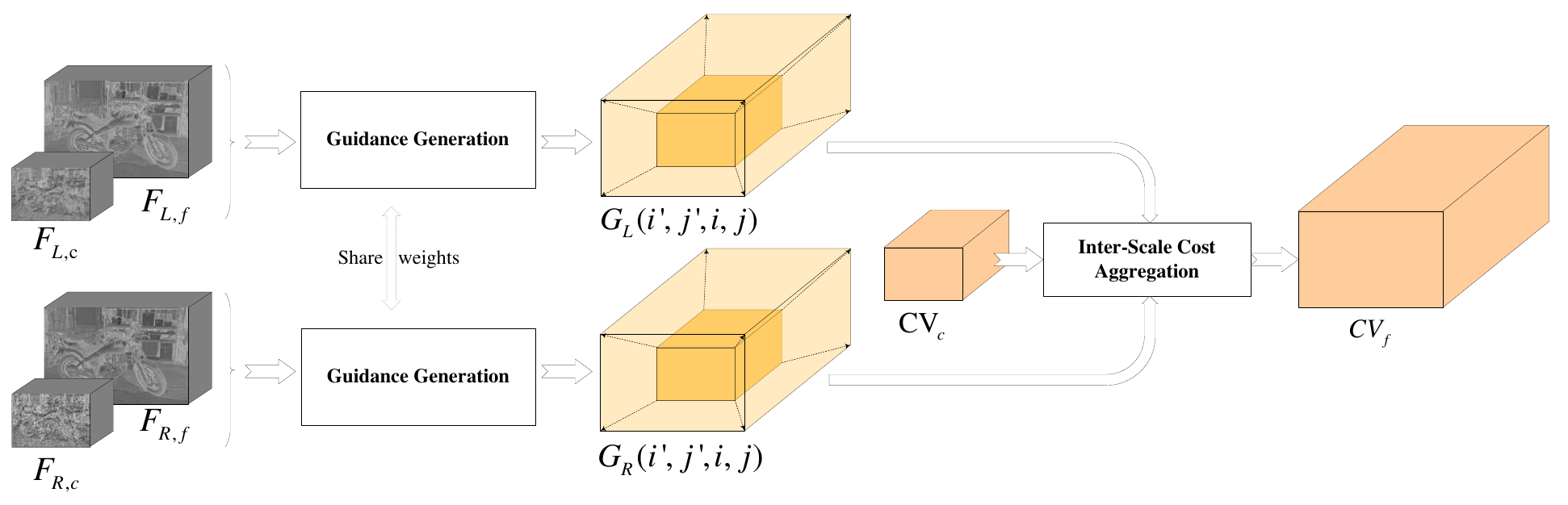}
        \caption{CAIS: Content-Aware Inter-Scale Cost Aggregation.}
        \end{minipage}%
    \end{subfigure}
	\centering
    \caption{Overview of our method.}
    \label{Fig:Model}
\end{figure}
\section{Method}
As shown in Figure \ref{Fig:Model}(b), our method includes a guidance generation module and an inter-scale cost aggregation module, which upsamples the cost volume for reliable detail recovery and aggregates cross-scale information to improve the quality of disparity estimation.

\subsection{Guidance Generation}
For content awareness, we embed our method in a stereo network, like PSMNet \cite{chang2018pyramid}, and extract the mapping between features from two scales as guidance where the features come from shallow layers of the network. We also design an explicit spatial encoding scheme, including the shifting of feature maps from two scales and the construction of location map, to learn guidance efficiently for following inter-scale cost aggregation.

\begin{figure}[htbp]
	\centering
	\includegraphics[width=.95\textwidth]{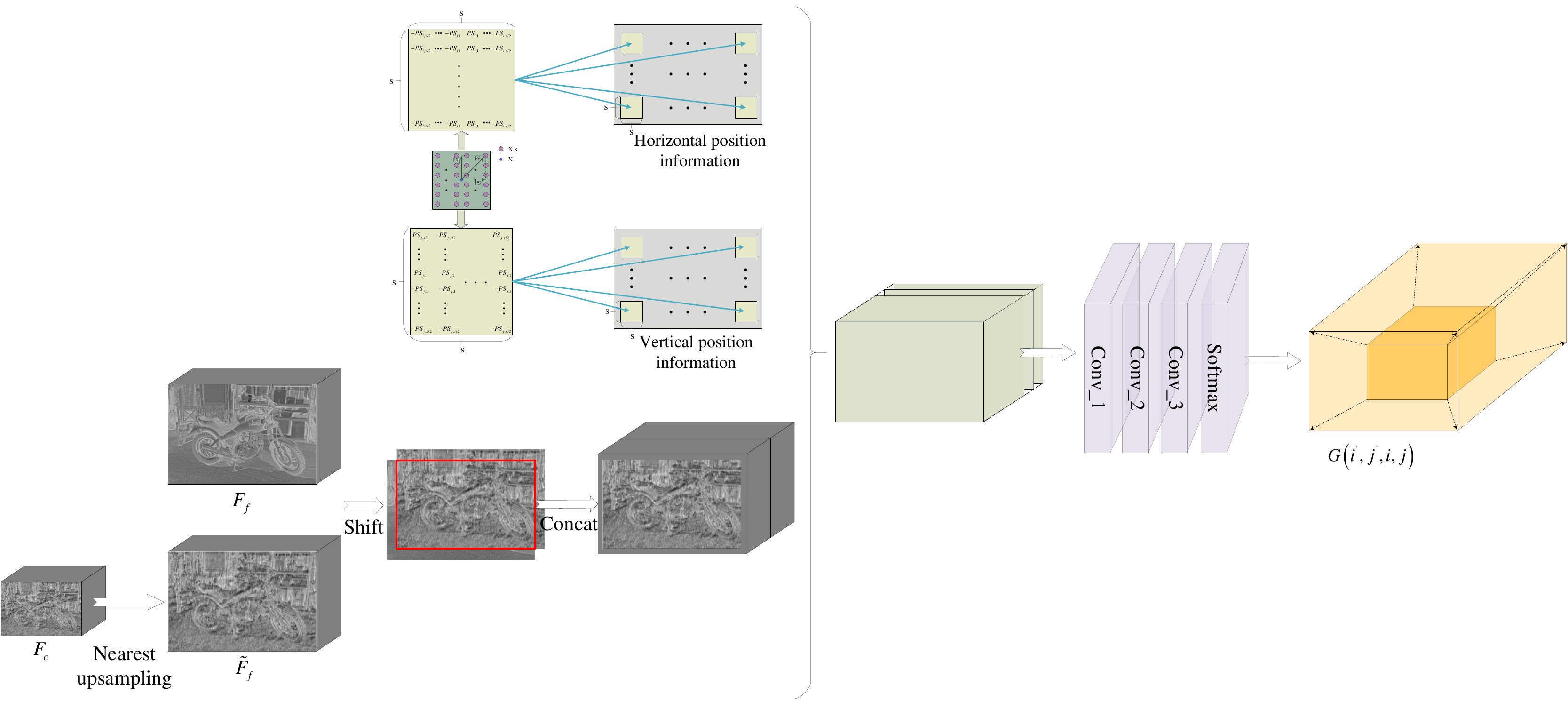}
	\centering
    \caption{Guidance generation module. We take the feature maps from two scales as input, and design an explicit spatial encoding scheme that includes the shifting of feature maps from two scales (the lower-left corner of the figure) and the construction of location maps (the upper-left corner of the figure). Then we use a series of $1 \times 1$ convolutions and softmax over the concatenation of feature maps and location maps to generate the guidance on the specific direction of shifting.}
    \label{Fig:SSM}
\end{figure}
Suppose there are feature map $F_{c}$ in coarse scale and feature map $F_{f}$ in fine scale where $c$ represents the coarse scale, $f$ represents the fine scale and the ratio between two scales is $s$. We first expand $F_{c}$ to $\widetilde{F}_{f}$ by nearest upsampling that is an identity mapping. Then, $F_{f}$ and $\widetilde{F}_{f}$ are concatenated after shifting with respect to the current spatial relationship. More specifically, we achieve the shifting via zero-padding to $F_{f}$ and $\widetilde{F}_{f}$ in opposite direction. For example, as shown in Figure \ref{Fig:SSM}, if the current window size is $3 \times 3$ and the target position of weight in the window is the upper-right corner, we then add a zero-padding with padding size of $s$ to the $F_{f}$ along upper-right and to the $\widetilde{F}_{f}$ along down-left. An adaptive location map $PS$ which will be explained in detail below is also concatenated to introduce the relative position information. The guidance $G_{dir}$ in direction $dir=(dir_{width}, dir_{height})$ are finally computed through three convolutions on the concatenation $F_{cat}$ :
\begin{equation}
   F_{cat} = [\widetilde{F}^{'}_{f}, F^{'}_{f}, PS].
\end{equation}
Taking ratio value $s\!=\!4$ and center direction where $dir\!=\!(0,0)$ as an example, $PS$ is designed as
\footnotesize
\[
PS_i \!=\! \operatorname{rp}\!\left( \begin{matrix}
   -2\!&\!-1\!&\!1\!&\! 2  \\
   -2\!&\!-1\!&\!1\!&\! 2  \\
   -2\!&\!-1\!&\!1\!&\! 2  \\
   -2\!&\!-1\!&\!1\!&\!2  \\
\end{matrix} \right),
PS_j \!=\! \operatorname{rp}\!\left( \begin{matrix}
   2\!&\!2\!&\!2\!&\!2  \\
   1\!&\!1\!&\!1\!&\!1  \\
   -1\!&\!-1\!&\!-1\!&\!-1  \\
   -2\!&\!-2\!&\!-2\!&\!-2 \\
\end{matrix} \right),
\]
\normalsize
\begin{equation}
    PS = [PS_i, PS_j],
\label{EQ:location map - center}
\end{equation}
where $rp(ele)$ is a process repeating $ele$ along rows and columns for $X$ times. Furthermore, the final guidance $G$ from different directions are regularized by softmax along direction dimension where $G(i',j',i,j)$ represents the mapping from position $(i,j)$ at coarse scale to position $(i',j')$ at fine scale. Then, we define the entire guidance generation module as $\mathcal{G}$:
\begin{equation}
\begin{aligned}
   G(i',j',i,j) &= \mathcal{G}(F_{f}(i',j'), F_{c}(i,j)), \\
   dir &= (i- \lfloor i'/s \rfloor, j- \lfloor j'/s \rfloor).
\end{aligned}
\label{EQ:guidance generation}
\end{equation}

\begin{figure}
  \begin{subfigure}{.35\textwidth}
    \centering
    \includegraphics[width=.95\textwidth]{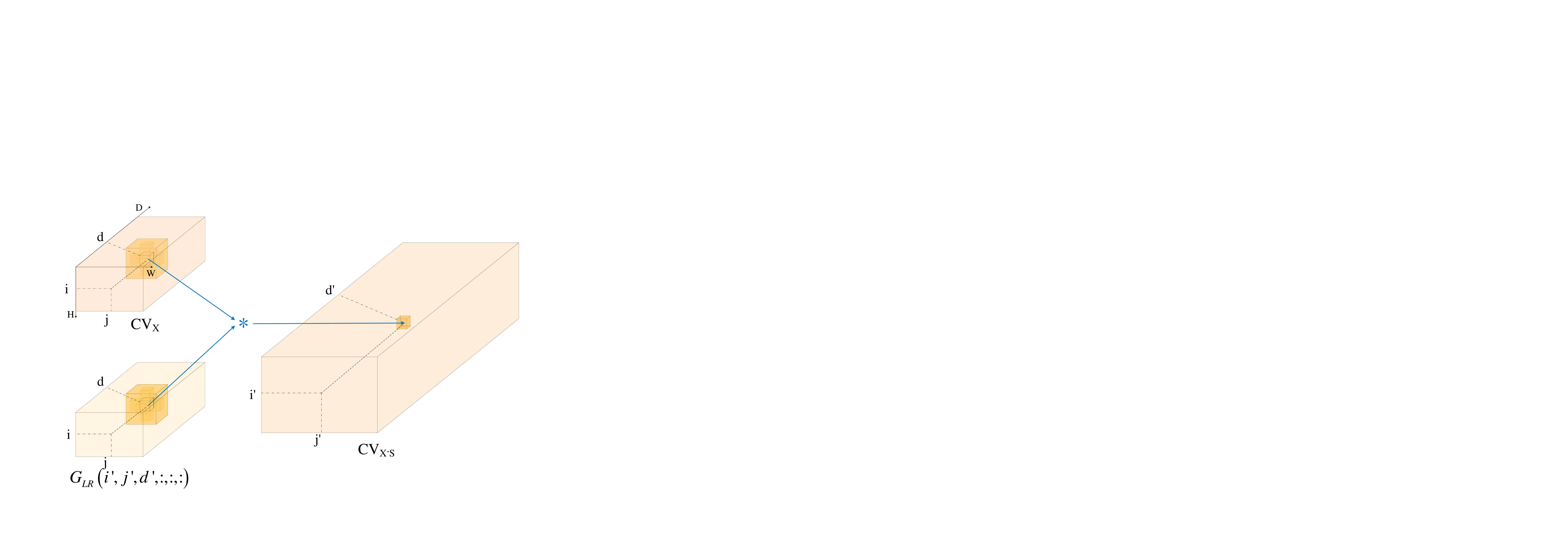}
    \caption{Full 3D upsampling.}
  \end{subfigure}
  \begin{subfigure}{.6\textwidth}
    \centering
    \includegraphics[width=.95\textwidth]{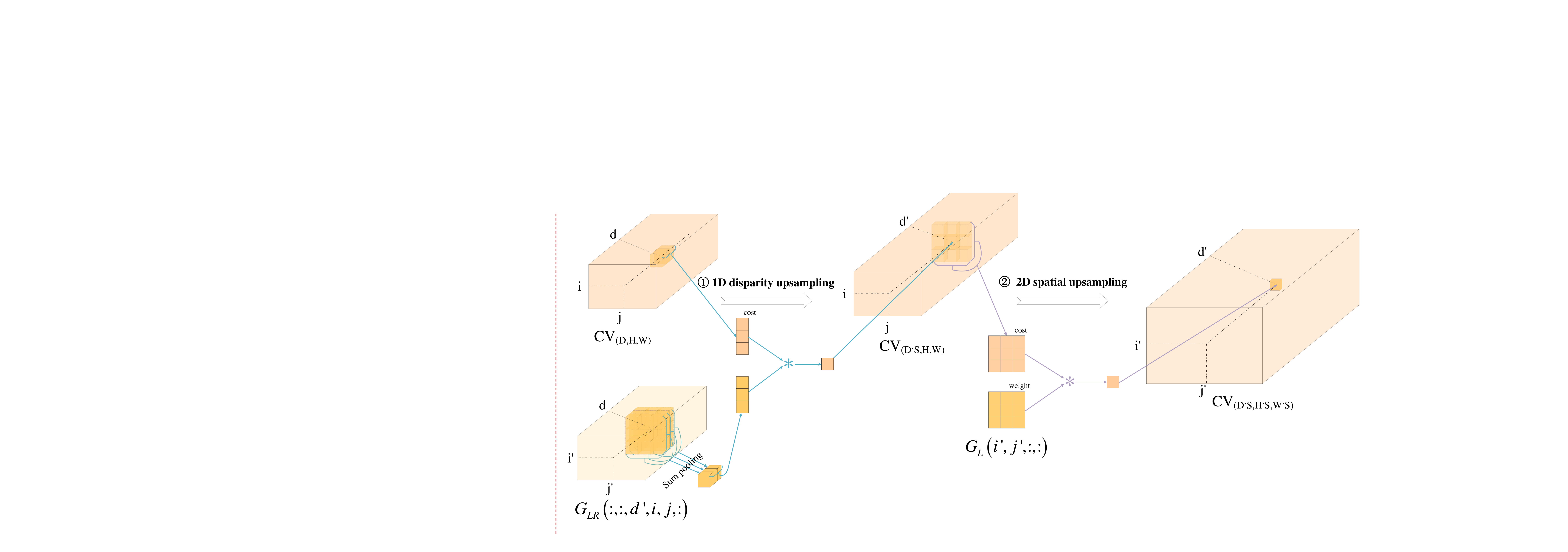}
    \caption{1D disparity/2D spatial upsampling.}
  \end{subfigure}
  \centering
    \caption{Illustration of inter-scale cost aggregation. We merge the guidance on left and right images into $G_{LR}$, and leave $G_L$ to represent the guidance generate from left image.}
    \label{Fig:cost_volume_upsampling}
\end{figure}

\subsection{Inter-Scale Cost Aggregation}
\subsubsection{Full 3D Spatial-Disparity Space}
In most stereo networks, the cost volume mapping is formulated as a weighted sum within the support window:
\begin{equation}
\begin{aligned}
   CV_{f}(x') = \sum_{x_n \in \mathcal{N}} W(x_n) \cdot CV_{c}(x_n),
\end{aligned}
\label{EQ:upsampling}
\end{equation}
where $x' \in [x \cdot s, (x+1) \cdot s)$, $x$ and $x_n$ are all 3D positions (spatial-disparity space) in cost volume, $CV_{c} \in \mathbb{R}^{W \times H \times D}$ and $CV_{f} \in \mathbb{R}^{W \cdot s \times H \cdot s \times D \cdot s}$ are the cost volume at coarse scale and fine scale respectively, $\mathcal{N}$ is the 3D window centered at $x$ and $W(x_n)$ is the weight of neighbor $x_n$.
We use feature maps from two scales to explicitly determine the weights in different areas, instead of applying the same weights over the entire cost volume (such as bilinear interpolation and deconvolution). We replace the weight $W(x)$ in Eq. \ref{EQ:upsampling} with the learned guidance $G_{LR}$ :
\begin{equation}
\begin{aligned}
    CV_{f}(i',j',d') = \sum_{(i,j,d) \in \mathcal{N}} &G_{LR}(i',j',d', i,j,d)) \\
     &\cdot CV_{c}(i,j,d),
\end{aligned}
\label{EQ:guided_aggregation}
\end{equation}
where the 3D position $x'$ and $x$ in Eq. \ref{EQ:upsampling} is unfold as $(i',j',d')$ and $(i,j,d)$. In deconvolution and bilinear interpolation, the $CV_{c}$ is filled with zeros to change its size before weighted summation, which means much zeros are appeared in the receptive field and provide no information. Differently, we directly sample from $CV_{c}$ which means, given a fixed size window, our method achieves much larger receptive field than deconvolution and bilinear interpolation.

As aforementioned, we do not directly learn the content-aware 3D kernel. Instead, we build the kernel on-the-spot from 2D guidance $G_L(i',j',i,j)$ and $G_R(i',j',i,j)$ ($L$: left view, $R$ right views):
\begin{equation}
\begin{aligned}
    G_{LR}(i',j',d', i,j,d) = G_L(i',j',i,j) \cdot G_R(i'-d',j',i-d,j),
\end{aligned}
\label{EQ:SLR}
\end{equation}
where $G_{L/R}$ is estimated based on the features $F_{L/R,f}$ and $F_{L/R,c}$, by a shared guidance generation module. We leverage the properties of cost volume in disparity dimension which corresponds to the pixel shift. In other words, the position $(i,j,d)$ in cost volume represents $(i,j)$ in left image and $(i-d, j)$ in right image. Through such a strategy, we improve the efficiency and reduce the computation cost compared to directly learn the 3D content-aware kernel.

Besides, we improve the efficiency by an additional space splitting scheme. In order to facilitate the explanation of this scheme, we rewrite the Eq. \ref{EQ:guidance generation} with Eq. \ref{EQ:guidance generation} and Eq. \ref{EQ:SLR} as 
\begin{equation}
\begin{aligned}
    CV_{(W,H,D) \cdot s}(i',j',d') = \sum_{(i,j,d) \in \mathcal{N}}
    CV_{(W,H,D)}(i,j,d) \cdot \\
    \mathcal{G}(F_{L,f}(i',j'), F_{L,c}(i,j)) \cdot \\
    \mathcal{G}(F_{R,f}(i'-d',j'), F_{R,c}(i-d,j)). \\
\end{aligned}
\label{EQ:guided_aggregation_on_cost_volume}
\end{equation}
Specifically, we split the full 3D spatial-disparity space into 1D disparity and 2D spatial space (see Figure \ref{Fig:cost_volume_upsampling}).

\subsubsection{1D Disparity Space}
In the first step which we call as 1D disparity inter-scale cost aggregation, the positions $(i,j,d-n), \dots, (i,j,d), \dots, (i,j,d+n)$ in cost volume along disparity dimension correspond to $(i,j)$ in left image and $(i-d+n,j), \dots, (i-d,j), \dots, (i-d-n,j)$ in right image, where $2n+1$ is the size of window $\mathcal{N}$. For the second step which we call as 2D spatial inter-scale cost aggregation, each $(i,j)$ in cost volume corresponds to $(i,j)$ in left image. More clearly, the 1D disparity inter-scale cost aggregation is formulated as
\begin{equation}
\begin{aligned}
    CV_{D \cdot s}(i,j,d') = \sum_{d} &\sum_{i'} \sum_{j'} \mathcal{G}(F_{R,f}(i'-d',j'),\\
    &F_{R,c}(i-d,j)) \cdot CV_{D}(i,j,d),
    \end{aligned}
\label{EQ:update along DD r}
\end{equation}

\begin{equation}
\begin{aligned}
    CV_{D \cdot s}(i,j,d') = \sum_{i'} \sum_{j'}
    &\mathcal{G}(F_{L, f}(i',j'), F_{L,c}(i,j))\\
    &\cdot CV_{D \cdot s}(i,j,d'),
\end{aligned}
\label{EQ:update along DD l}
\end{equation}
where $i' \in [i \cdot s, (i+1) \cdot s)$, $j' \in [j \cdot s, (j+1) \cdot s)$, $d \in \{ \lfloor d'/s \rfloor-1, \lfloor d'/s \rfloor, \lfloor d'/s \rfloor+1 \}$, $F_L$ and $F_R$ are the feature maps respectively from left and right views. 
\subsubsection{2D Spatial Space}
As to the 2D spatial inter-scale cost aggregation, it is formulated as
\begin{equation}
\begin{aligned}
    CV_{(W,H) \cdot s}(i',j',:) = \! \sum_{(i,j) \in \mathcal{N}_{sp}} \! &\mathcal{G}(F_{L, f}(i',j'), F_{L, c}(i,j))\\
    &\cdot CV_{(W,H)}(i,j,:),
\end{aligned}
\label{EQ:update along SD}
\end{equation}
where $\mathcal{N}_{sp} = \{ \lfloor i'/s \rfloor-1, \lfloor i'/s \rfloor, \lfloor i'/s \rfloor+1 \} \times \{ \lfloor j'/s \rfloor-1, \lfloor j'/s \rfloor, \lfloor j'/s \rfloor+1 \}$. Through the above two-step splitting scheme, we achieve the transformation from $W \times H \times D$ to $W \times H \times D \cdot s$ and then from $W \times H \times D \cdot s$ to $W \cdot s \times H \cdot s \times D \cdot s$. We also greatly reduce the computation complexity in terms of FLOPs (floating-point operations) even compared to 3D de-convolution.

\section{Experiments}
In this section, we evaluate our method on a synthetic dataset (Scene Flow \cite{guney2015displets}), an outdoor dataset (KITTI 2015 \cite{menze2015object}), and an indoor dataset (Middlebury-v3 \cite{scharstein2014high}). We use baseline network PSMNet \cite{chang2018pyramid} to demonstrate the effectiveness of our method, by applying our method to replace the bilinear interpolation and deconvolution in the PSMNet. All models are implemented via PyTorch and optimized using the Adam algorithm with $\beta_1$ of 0.9, $\beta_2$ of 0.999 and batch size of 4. We use four NVIDIA GTX 1080Ti GPUs for training. During training, color normalization is applied to each input image which is then cropped to $256 \times 512$ resolution. We train our network on Scene Flow for 10 epochs with learning rate of 0.001. We then fine-tune the network on KITTI 2015 and set the learning rate to 0.001, 0.0001 and 0.00003 for the first 200 epochs, the next 400 epochs and an additional 600 epochs respectively. As for Middlebury-v3, we also fine-tune the model pre-trained on Scene Flow. The learning rate is set to 0.001 for 300 epochs and then changed to 0.0001 for the rest 600 epochs. In all the experiments, no post-processing or unsupervised learning is used.

\begin{table}[htbp]
  \centering
  \caption{The ablation study of the superiority of our content-aware inter-scale cost aggregation. A $\rightarrow$ B represents the replacement of A with B.}
  \scalebox{1.}[1.1]{%
  \setlength{\tabcolsep}{2mm}{%
    \begin{tabular}{c|c|c|c}
    \hline
    Models &  \tabincell{c}{Replacing Bilinear\\Upsampling} & Replacing deconv & EPE \\
    \hline
    \hline
    PSMNet \cite{chang2018pyramid} & & &  1.09 \\
    \hline
    PSMNet (Bilinear $\rightarrow$ Ours) & \checkmark & &  0.61 \\
    \hline
    PSMNet (Deconv $\rightarrow$ Ours) & & \checkmark & 0.76 \\
    \hline
    PSMNet (All $\rightarrow$ Ours) & \checkmark & \checkmark &  0.57 \\
    \hline
    \hline
    \end{tabular}%
  }
  }
  \label{tab:ablation}%
\end{table}%

\begin{table}[htbp]
  \centering
  \caption{The ablation study of the effectiveness of our strategy for content-aware weight learning.}
  \scalebox{1.}[1.1]{%
  \setlength{\tabcolsep}{2mm}{%
    \begin{tabular}{c|c}
    \hline
    Models & EPE \\
    \hline
    \hline
    Full &  0.57 \\
    \hline
    w/o Spatial Relationship Encoding &  0.67 \\
    \hline
    w/o Stereo Views & 0.63 \\
    \hline
    \end{tabular}%
  }
  }
  \label{tab:ablation2}%
\end{table}%

\subsection{Ablation Study}
An ablation study is conducted on the Scene Flow dataset to show the benefits of each sub-parts of our methods. We test the superiority of our method over bilinear interpolation and deconvolution. The effectiveness of our strategy for content-aware weight learning is verified by identifying the importance of explicit spatial encoding and the use of stereo views for weight learning. We also test a more challenging situation to demonstrate the power of our method.

\subsubsection{Superiority Compared to Bilinear Interpolation}
In order to show the superiority of our method over bilinear interpolation, we compare the result of the PSMNet with ours that is generated by replacing the bilinear interpolation in the PSMNet with our method. As depicted in Table \ref{tab:ablation}, the PSMNet (Bilinear $\rightarrow$ ours) improve the result of the PSMNet by 0.48 in EPE, which shows the superiority of proposed method over bilinear interpolation for cost volume upsampling.

\subsubsection{Superiority Compared to Deconvolution}
We also analyze the superiority of our method when compared to deconvolution. We replace the two deconvolutions in the PSMNet with our method. As shown in Table \ref{tab:ablation}, the PSMNet (Deconv $ \ rightarrow $ ours) outperforms the PSMNet by 0.33 in EPE,  which shows the superiority of proposed method over deconvolution for inter-scale cost aggregation. We further test the performance of the PSMNet (All $\rightarrow$ ours) which is used in the following benchmark performance comparison. Additional improvement can be observed when both bilinear interpolation and deconvolution are replaced.

\subsubsection{Importance of Explicit Spatial Relationship Encoding}
We verify the importance of using explicit spatial relationship encoding to learn the content-aware similarity/affinity weights. As shown in Table \ref{tab:ablation2}, the performance drops sharply if we use the feature input by a direct concatenation of feature maps from two scales without explicit spatial encoding. This situation can be viewed as unitary feature input by fusion of multi-scale feature maps. 

\subsubsection{Importance of Stereo Views}
We verify the importance of using 3D information from stereo views to learn content-aware weight for the aggregation on 3D cost volume. As presented in Table \ref{tab:ablation2}, worse result is obtained by using the guidance based on only 2D textural information from the left view.

\subsubsection{Performance in Challenging Situation}
In addition to the above experiments, we test the performance of our method in a more challenging situation. Instead of upsampling the cost volume from $1/8$ to $1/4$ and then $1/4$ to $1$, we directly transform the cost volume from ratio $1/8$ to $1$. In this challenging situation, our method achieves accuracy of 0.89 in EPE which is still better than the result of PSMNet. Such results not only demonstrate the power of our method but also indicate its potential applications under resource constrained situations like mobile phones.



\subsection{Benchmark Performance}

\begin{table}[htbp]
  \centering
  \caption{Evaluation on Scene Flow dataset.}
  \scalebox{1.}[1.1]{%
  \setlength{\tabcolsep}{3mm}{%
    \begin{tabular}{cc||cc}
    \hline
    Models  & EPE   & Models  & EPE  \\
    \hline
    \hline
    DispNetC \cite{mayer2016large} & 1.68  & SSPCV-Net \cite{wu2019semantic} & 0.87  \\
    CRL \cite{pang2017cascade} & 1.32  & DeepPruner-best \cite{duggal2019deeppruner} & 0.86  \\
    GCNet \cite{kendall2017end} & 1.84  & GA-Net-15 \cite{zhang2019ga} & 0.84  \\
    PDS-Net \cite{tulyakov2018practical} & 1.12  & CSPN \cite{cheng2019learning} & 0.78 \\
    EdgeStereo \cite{song2018edgestereo} & 1.11  & GwcNet-gc \cite{guo2019group} & 0.77 \\
    StereoNet \cite{khamis2018stereonet} & 1.10  &  EdgeStereo V2 \cite{song2019edgestereo} & 0.74 \\
    PSMNet \cite{chang2018pyramid} & 1.09  &  Ours  & \textbf{0.57} \\
    \hline
    \end{tabular}%
  }
  }
  \label{tab:Scene_Flow}%
\end{table}%

\begin{figure}[htbp]
	\centering
	\includegraphics[width=.98\textwidth]{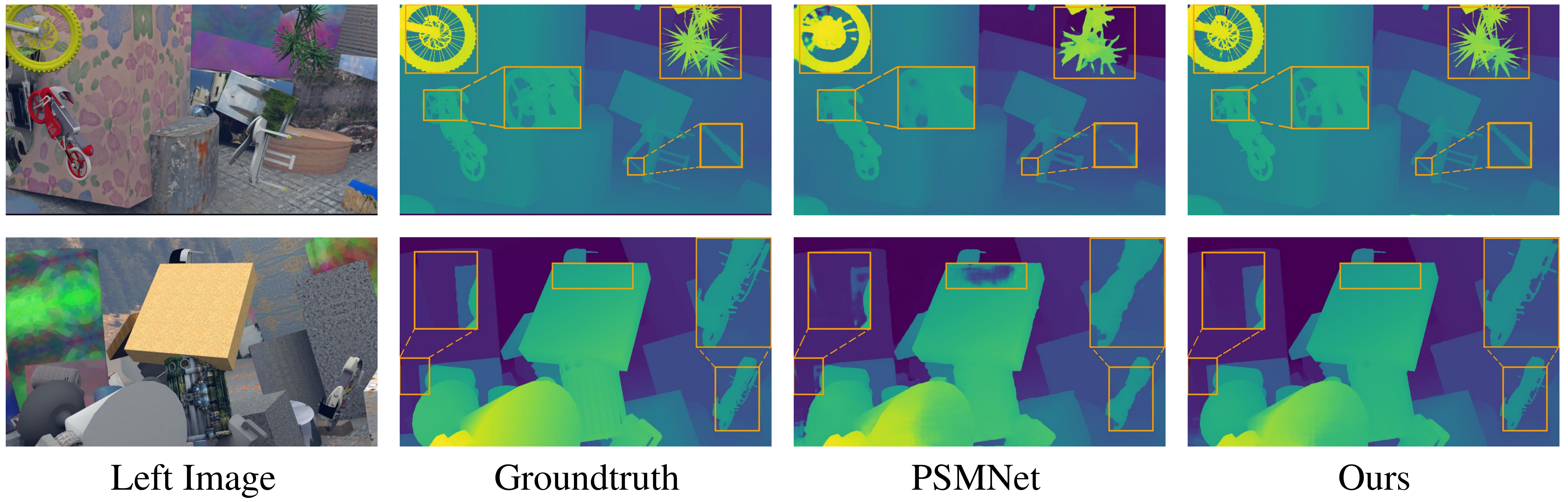}
	\centering
    \caption{The visualization of results on KITTI 2015 Scene Flow dataset. The first column is the left input image. The second column is the corresponding ground truth. The third and last column is the results of PSMNet   and ours, respectively.}
    \label{Fig:EXP_Scene Flow benchmark}
\end{figure}
\subsubsection{Scene Flow Dataset}
Scene Flow is a large synthetic dataset containing 34,896 training images and 4,248 testing images with size of $540 \times 960$. This dataset has three rendered sub-datasets: FlyingThings3D, Monkaa, and Driving. FlyingThings3D is rendered from the ShapeNet dataset and has 21,828 training data and 4248 testing data. Monkaa is rendered from the animated film Monkaa and has 8666 training data. The Driving is constructed by the naturalistic, dynamic street scene from the viewpoint of a driving car and has 4,402 training samples.

We compare our model with other state-of-the-art methods on this dataset. As shown in Table \ref{tab:Scene_Flow}, our model achieve the best end-point-error (EPE) 0.57, almost twice better than the baseline, PSMNet. We also visualize the results of PSMNet and ours. In Fig. \ref{Fig:EXP_Scene Flow benchmark}, we can easily observe the significant improvements of our method not only in the fine-grained areas but also in the large textureless areas. For 

\begin{figure*}[htbp]
	\centering
	\includegraphics[width=.98\textwidth]{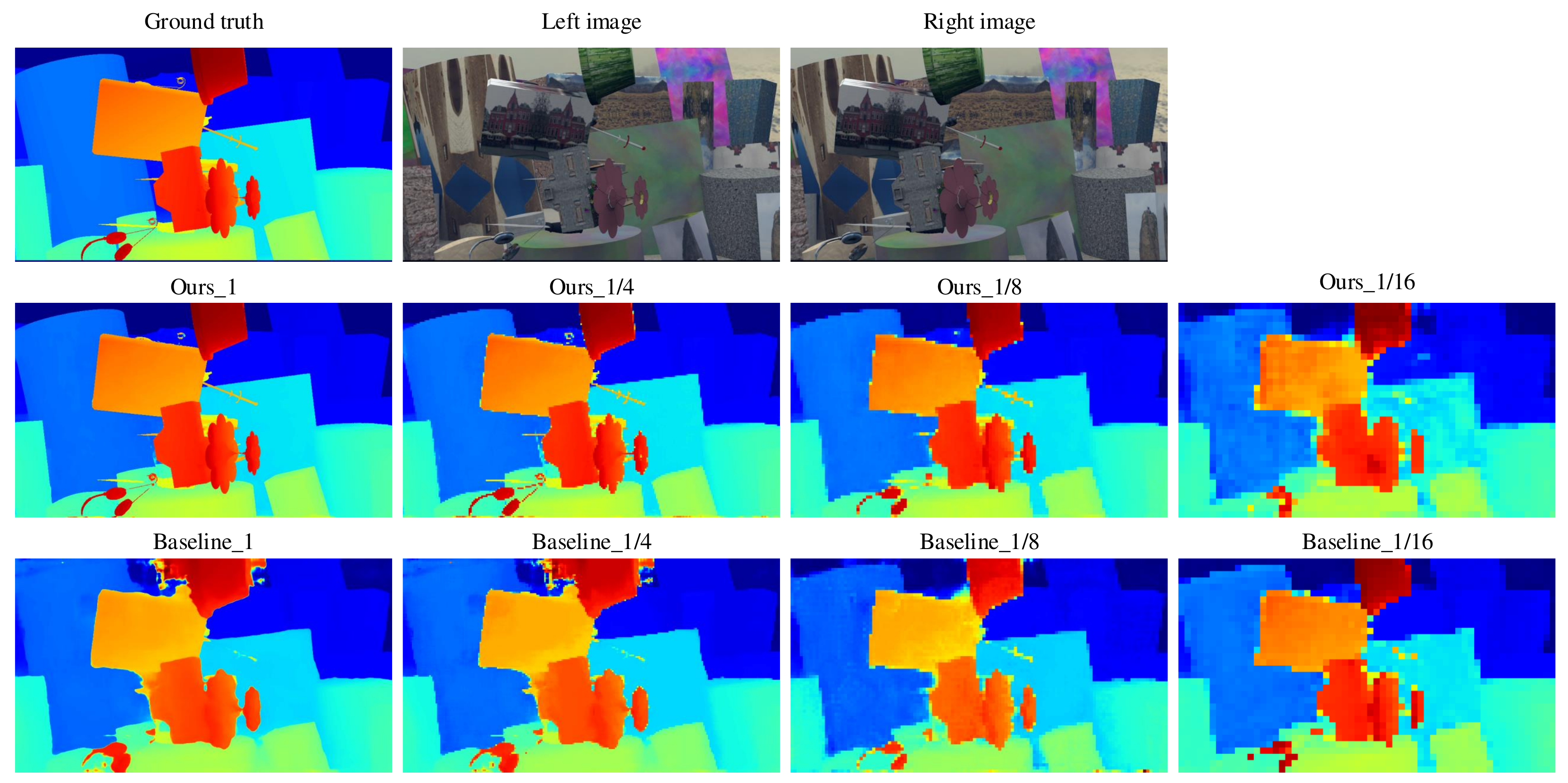}
	\centering
    \caption{The visualization of multi-scale outputs of ours and the baseline model. Ours/Baseline\_1, Ours/Baseline\_1/4, Ours/Baseline\_1/8 and Ours/Baseline\_1/16 represent the disparity maps with scale 1, 1/4, 1/8 and 1/16 respectively. Here, for better visualization, we zoom in the disparity maps at the scale 1/4, 1/8 and 1/16.}
    \label{Fig:3188}
\end{figure*}

\begin{figure*}[htbp]
	\centering
	\includegraphics[width=.98\textwidth]{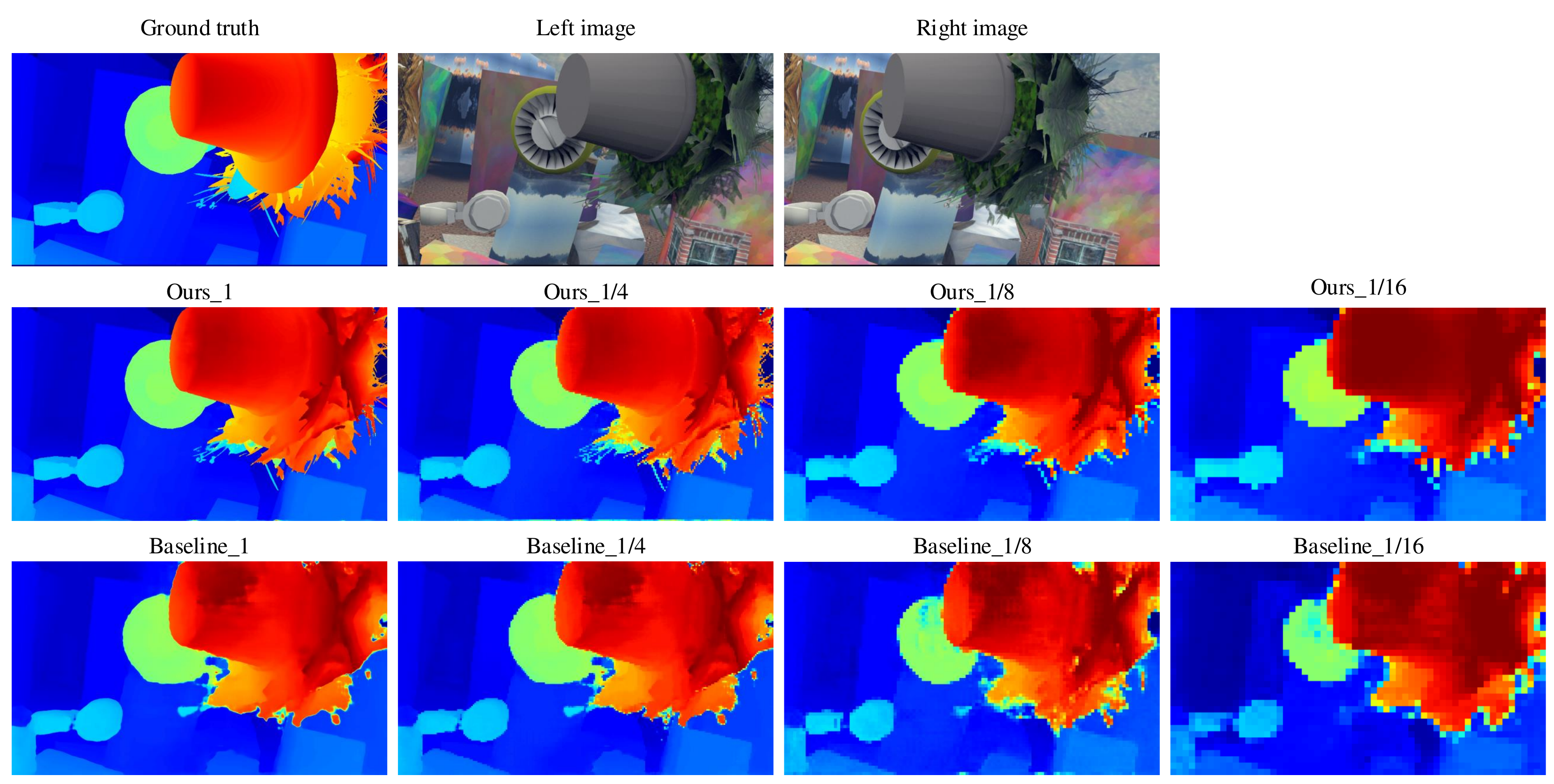}
	\centering
    \caption{The visualization of multi-scale outputs of ours and the baseline model. Ours/Baseline\_1, Ours/Baseline\_1/4, Ours/Baseline\_1/8 and Ours/Baseline\_1/16 represent the disparity maps with scale 1, 1/4, 1/8 and 1/16 respectively. Here, for better visualization, we zoom in the disparity maps at the scale 1/4, 1/8 and 1/16.}
    \label{Fig:1632}
\end{figure*}

\begin{figure*}[htbp]
	\centering
	\includegraphics[width=.98\textwidth]{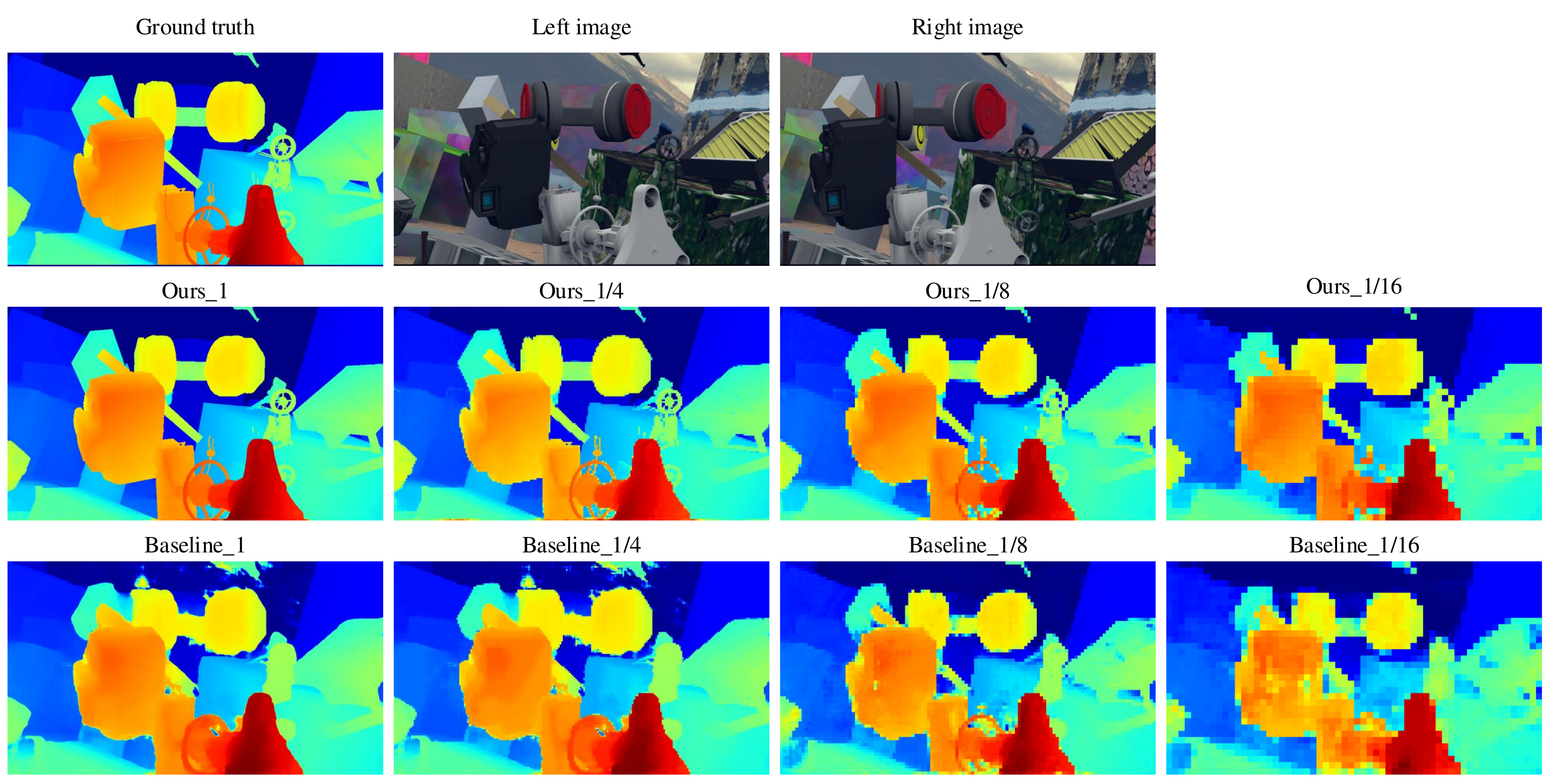}
	\centering
    \caption{The visualization of multi-scale outputs of ours and the baseline model. Ours/Baseline\_1, Ours/Baseline\_1/4, Ours/Baseline\_1/8 and Ours/Baseline\_1/16 represent the disparity maps with scale 1, 1/4, 1/8 and 1/16 respectively. Here, for better visualization, we zoom in the disparity maps at the scale 1/4, 1/8 and 1/16.}
    \label{Fig:53}
\end{figure*}

\begin{figure*}[htbp]
	\centering
	\includegraphics[width=.98\textwidth]{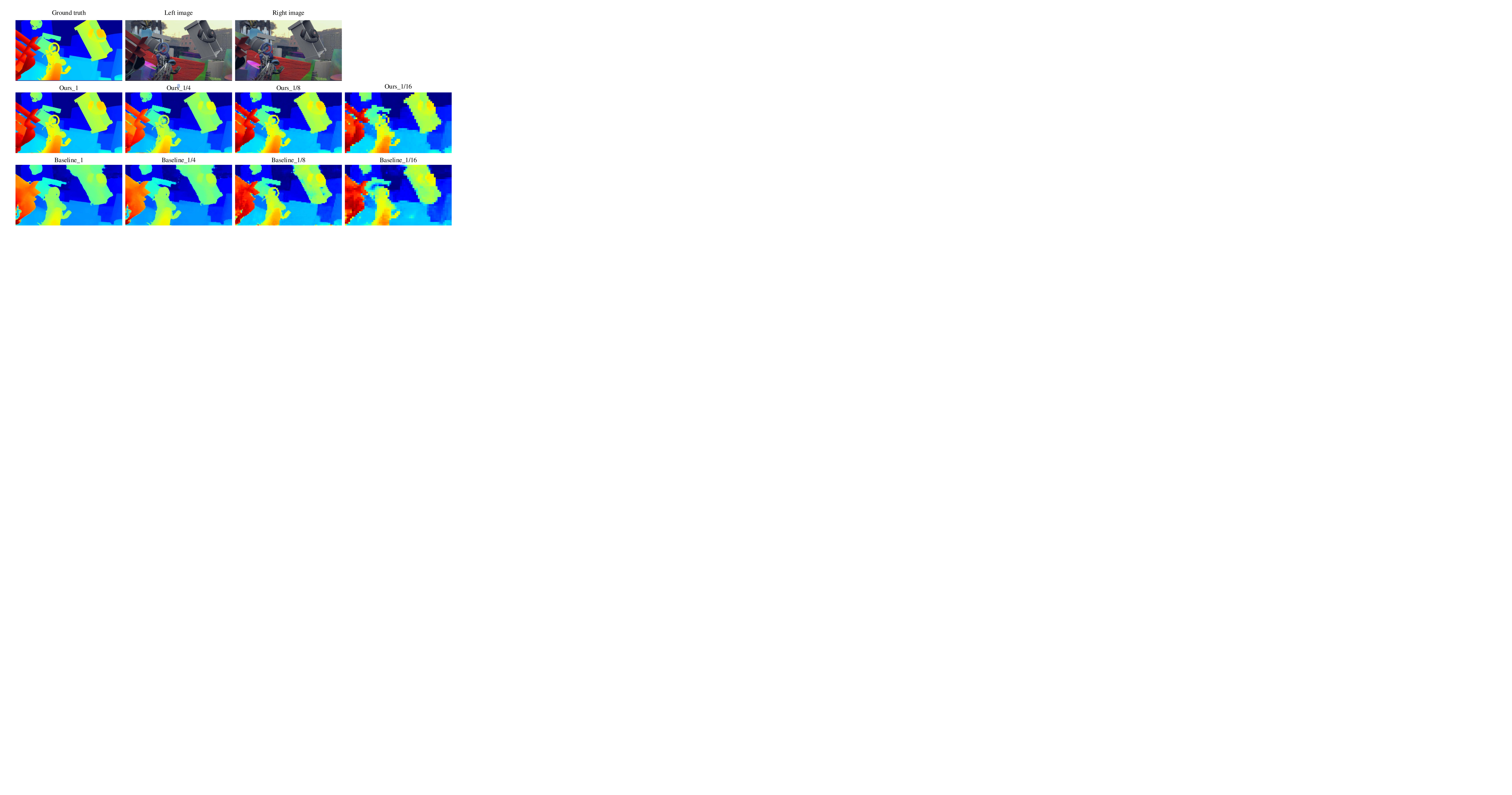}
	\centering
    \caption{The visualization of multi-scale outputs of ours and the baseline model. Ours/Baseline\_1, Ours/Baseline\_1/4, Ours/Baseline\_1/8 and Ours/Baseline\_1/16 represent the disparity maps with scale 1, 1/4, 1/8 and 1/16 respectively. Here, for better visualization, we zoom in the disparity maps at the scale 1/4, 1/8 and 1/16.}
    \label{Fig:2081}
\end{figure*}

\begin{figure*}[htbp]
	\centering
	\includegraphics[width=.98\textwidth]{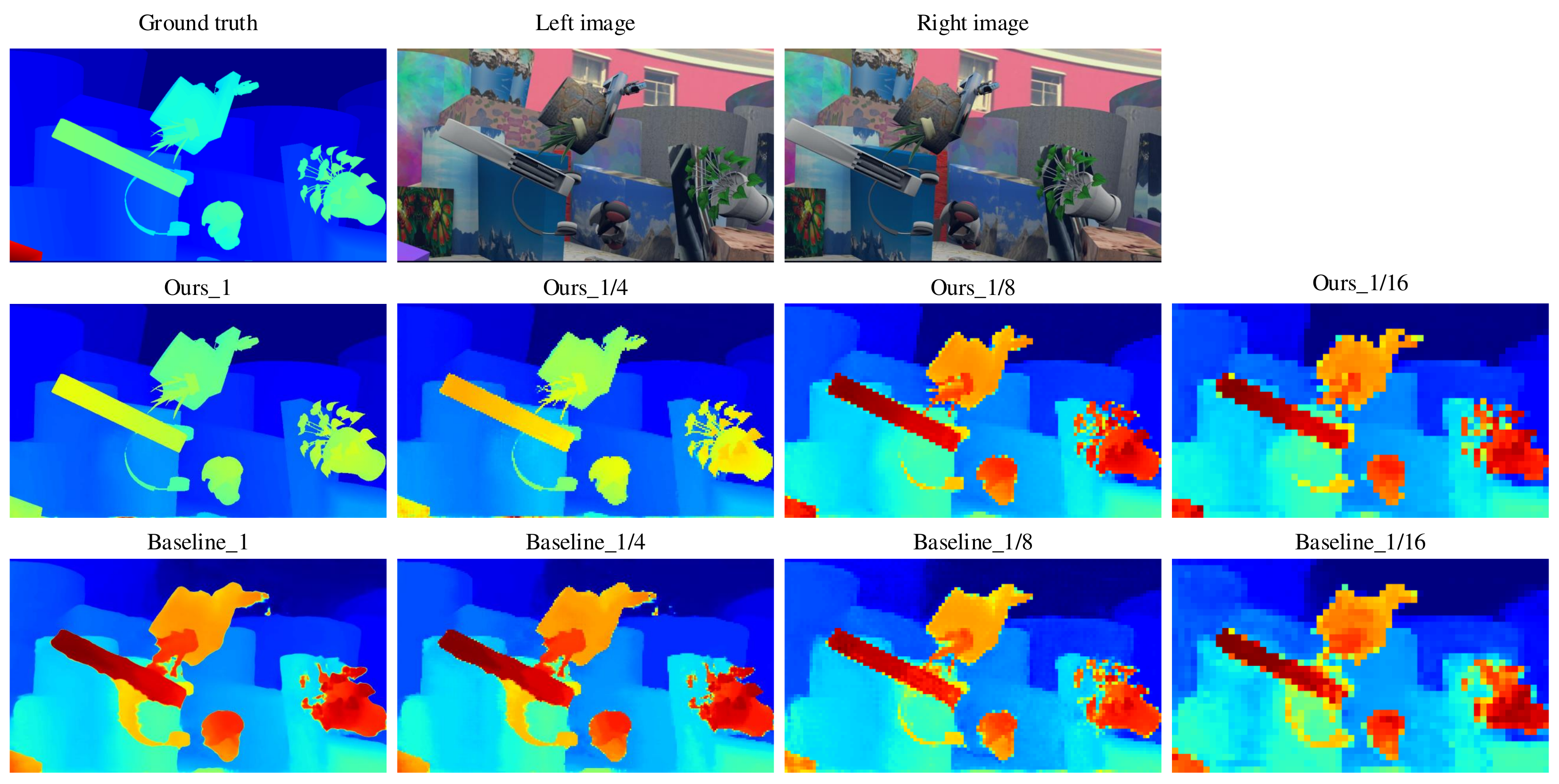}
	\centering
    \caption{The visualization of multi-scale outputs of ours and the baseline model. Ours/Baseline\_1, Ours/Baseline\_1/4, Ours/Baseline\_1/8 and Ours/Baseline\_1/16 represent the disparity maps with scale 1, 1/4, 1/8 and 1/16 respectively. Here, for better visualization, we zoom in the disparity maps at the scale 1/4, 1/8 and 1/16.}
    \label{Fig:3695}
\end{figure*}

\begin{figure*}[htbp]
	\centering
	\includegraphics[width=.98\textwidth]{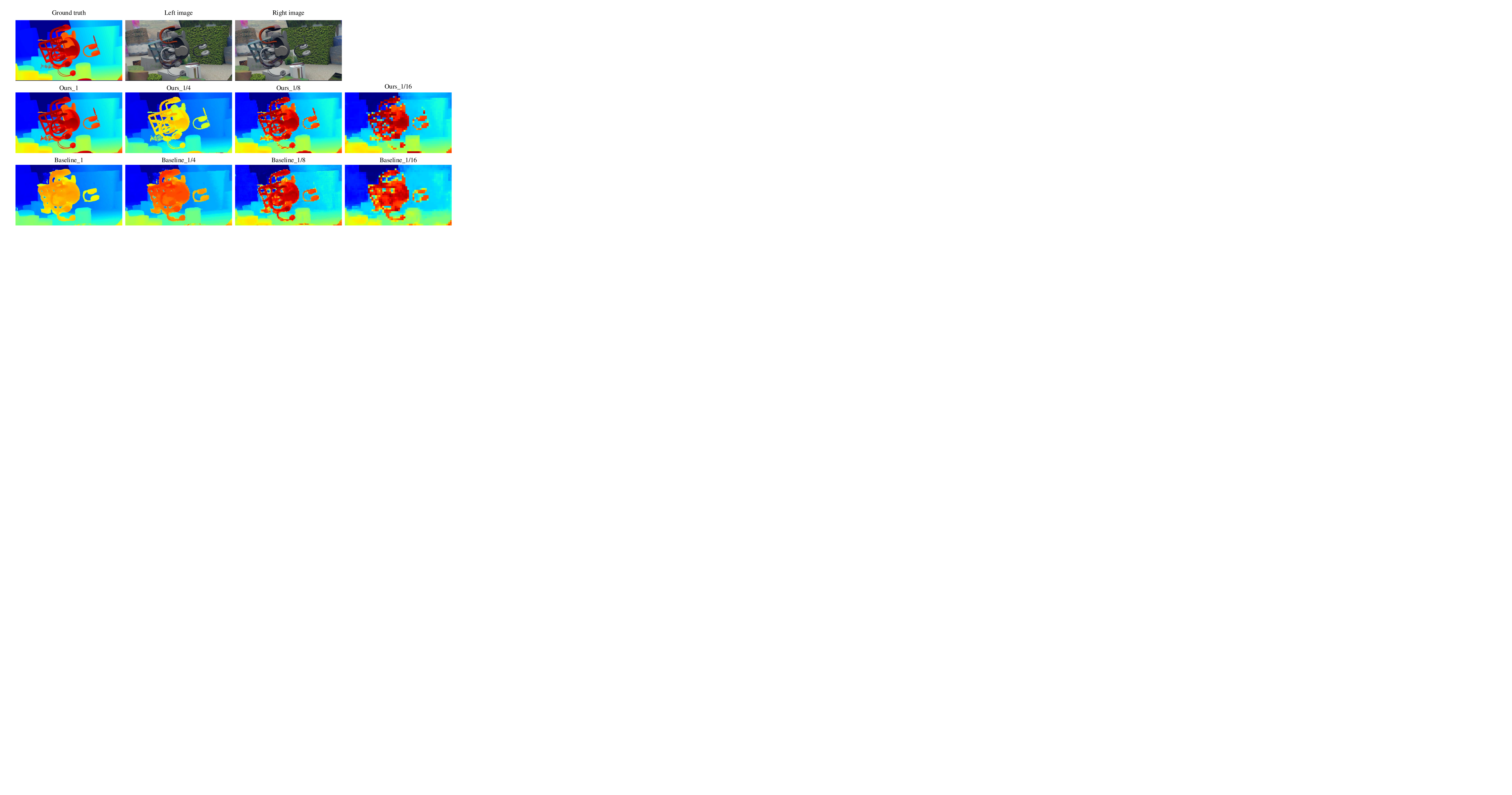}
	\centering
    \caption{The visualization of multi-scale outputs of ours and the baseline model. Ours/Baseline\_1, Ours/Baseline\_1/4, Ours/Baseline\_1/8 and Ours/Baseline\_1/16 represent the disparity maps with scale 1, 1/4, 1/8 and 1/16 respectively. Here, for better visualization, we zoom in the disparity maps at the scale 1/4, 1/8 and 1/16.}
    \label{Fig:3424}
\end{figure*}

In order to better visualize the capability of our method in detail recovery, we additionally supply a set of multi-scale disparity maps that are obtained from our method and PSMNet (see Figure \ref{Fig:3188} - \ref{Fig:3424}). As the PSMNet only outputs the full-resolution result, we add the same regression model onto each cost volume at different scales, and retrain it on Sceneflow. From the multi-scale illustration (Figure \ref{Fig:3188} - \ref{Fig:3424}), it is obviously that our method avoids overlapping the foreground and background, and achieve reliable recovery of details which are important for the high-quality 3D reconstruction and the detection of distant objects in autonomous driving.

\begin{table}[htbp]
  \centering
  \caption{Evaluation on KITTI 2015 Benchmark}
  \scalebox{1.}[1.1]{%
  \setlength{\tabcolsep}{2mm}{%
    \begin{tabular}{c|ccc|ccc|c}
    \hline
    \multirow{2}[4]{*}{Models} & \multicolumn{3}{c|}{Noc (\%)} & \multicolumn{3}{c|}{All (\%)} & \multirow{2}[4]{*}{time (s)} \\
\cline{2-7}          & \multicolumn{1}{c|}{bg} & \multicolumn{1}{c|}{fg} & all   & \multicolumn{1}{c|}{bg} & \multicolumn{1}{c|}{fg} & all   &  \\
    \hline
    \hline
    SGM-Nets \cite{seki2017sgm} & 2.23  & 7.43  & 3.09  & 2.66  & 8.64  & 3.66  & 67 \\
    GCNet \cite{kendall2017end} & 2.02  & 5.58  & 2.61  & 2.21  & 6.16  & 2.87  & 0.90 \\
    PDS-Net \cite{tulyakov2018practical} & 2.09  & 3.68  & 2.36  & 2.29  & 4.05  & 2.58  & 0.50 \\
    EdgeStereo \cite{song2018edgestereo} & 2.12  & 3.85  & 2.40  & 2.27  & 4.18  & 2.59  & 0.27 \\
    StereoNet \cite{khamis2018stereonet} & 4.05  & 6.44  & 4.44  & 4.3   & 7.45  & 4.83  & 0.02 \\
    PSMNet \cite{chang2018pyramid} & 1.71  & 4.31  & 2.14  & 1.86  & 4.62  & 2.32  & 0.41 \\
    SSPCV-Net \cite{wu2019semantic} & 1.61  & 3.40  & 1.91  & 1.75  & 3.89  & 2.11  & 0.90 \\
    GwcNet-g \cite{guo2019group} & 1.61  & 3.49  & 1.92  & 1.74  & 3.93  & 2.11  & 0.32 \\
    EMCUA \cite{nie2019multi} & 1.50  & 3.88  & 1.90  & 1.66  & 4.27  & 2.09  & 0.32 \\
    DeepPruner-best \cite{duggal2019deeppruner} & 1.71  & \textbf{3.18}  & 1.95  & 1.87  & \textbf{3.56}  & 2.15  & 0.18 \\
    EdgeStereo V2 \cite{song2019edgestereo} & 1.69 & \textbf{2.94} & 1.89 & 1.84 & \textbf{3.30} & 2.08 & 0.32 \\
    GANet-15 \cite{zhang2019ga} &  1.40  & 3.37  & 1.73  &  \textbf{1.55}  & 3.82  & 1.93 & 0.36 \\
    CSPN \cite{cheng2019learning} & 1.32  & \textbf{2.87}  & \textbf{1.58}  & \textbf{1.51}  & \textbf{2.88}  & \textbf{1.74}  & 1.00 \\
    Ours  & \textbf{1.29}  & \textbf{3.29}  & \textbf{1.70}  & \textbf{1.57}  & \textbf{3.62}  & \textbf{1.91}  & \textbf{0.38} \\
    \hline
    \end{tabular}%
  }
  }
  \label{tab:KITTI15}%
\end{table}%

\begin{figure*}[htbp]
	\centering
	\includegraphics[width=.98\textwidth]{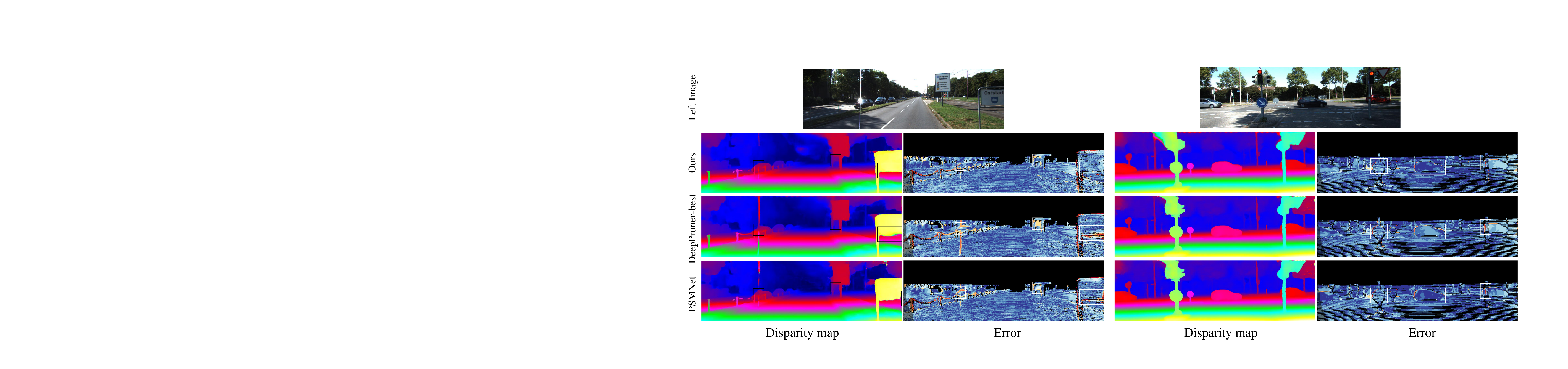}
	\centering
    \caption{The visualization of results on KITTI 2015. The first row is the left input image. The second row is result of ours. The third and last row are the results of DeepPruner-best and PSMNet, respectively. The first and third columns are disparity maps. The second and fourth columns are corresponding D1 error maps.}
    \label{Fig:KITTI15}
\end{figure*}

\subsubsection{KITTI 2015 Dataset}
Unlike the synthetic dataset SceneFlow, KITTI 2015 is a real-world dataset with street views from a driving car. It contains 200 training stereo image pairs with sparse ground-truth disparities obtained using LiDAR and another 200 testing image pairs without ground-truth disparities. We took 160 images for training and left 40 images for evaluation.

In order to show the performance of our model in a real outdoor scene, we test it on KITTI 2015 and compare it with state-of-the-art methods. As depicted in Table \ref{tab:KITTI15}, we achieve competitive performance on KITTI 2015 benchmark. Although CSPN \cite{cheng2019learning} is around $0.0012$ better than ours in Noc-all and $0.0017$ better than ours in All-all, the speed of our method is nearly 3 times faster than theirs. Furthermore, CSPN uses Nvidia P40 while our model are benchmarked on GTX 1080Ti. Apart from the quantitative analysis, a qualitative comparison between our model and PSMNet  is also given in Figure \ref{Fig:KITTI15}. The differences between PSMNet  and ours can be easily found in the above figures, especially in the black and white boxes we marked.

\begin{table*}[htbp]
  \centering
  \caption{Evaluation on Middlebury-v3. For comparison, we mainly focus on the methods with quarter (Q) resolution.}
  \scalebox{1.}[1.1]{%
  \setlength{\tabcolsep}{2mm}{
    \begin{tabular}{c|c|cccccc|c}
    \hline
    Models & Res & bad 1 & bad 4 & avgerr & rms   & A90   & A95   & time \\
    \hline
    \hline
    MotionStereo \cite{valentin2018depth} & H & 58.4 & 31.4 & 19.2  & 41.6  & 63.2  & 97.3  & 0.1 \\
    PDS \cite{tulyakov2018practical}  & H & \textbf{38.3} & \textbf{6.98} & 6.9   & 27.5  & \textbf{9.95}  & \textbf{35.5}  & 12.5 \\
    iResNet\_ROB \cite{liang2018learning} & H & 45.9 & 15.8 & \textbf{6.56}  & \textbf{18.1}  & 15.1  & 36.2  & 0.34 \\
    \hline
    DDL \cite{yin2017sparse} & Q & 63.9 & 13.6 & 10.8 & 32.6 & 30.2 & 71.3 & 112 \\
    IGF \cite{hamzah2017stereo} & Q & 66.2 & 18 & 11 & 31 & 30.6 & 65.7 & 132 \\
    DSGCA \cite{park2018deep} & Q & 67.5 & 18 & 26.9 & 66.6 & 102 & 173 & 11 \\
    MSMD\_ROB \cite{lu2018cascaded} & Q & 62.3 & 15.3 & 17.4 & 50.4 & 62.5 & 127 & 0.79 \\
    PSMNet\_ROB \cite{chang2018pyramid} & Q & 67.3 & 23.5 & 8.78  & 23.3  & 22.8  & 43.4 & 0.64 \\
    AMNet \cite{du2019amnet} & Q & 77.5 & 19.2 & 7.58 & 22.9 & 16.1 & 35.6 & 0.4 \\
    DeepPruner\_ROB \cite{duggal2019deeppruner} & Q & \textbf{57.1} & 15.9 & 6.56 & 18 & 17.9 & 33.1 & \textbf{0.13} \\
    Ours  & Q & \textbf{59.7} & \textbf{13.7} & \textbf{5.43} & \textbf{17.3} & \textbf{8.11} & \textbf{25.2} & \textbf{0.28} \\
    \hline
    \end{tabular}%
  }
  }
  \label{tab:Middlebury-v3}%
\end{table*}%

\begin{figure*}[htbp]
	\centering
	\includegraphics[width=.98\textwidth]{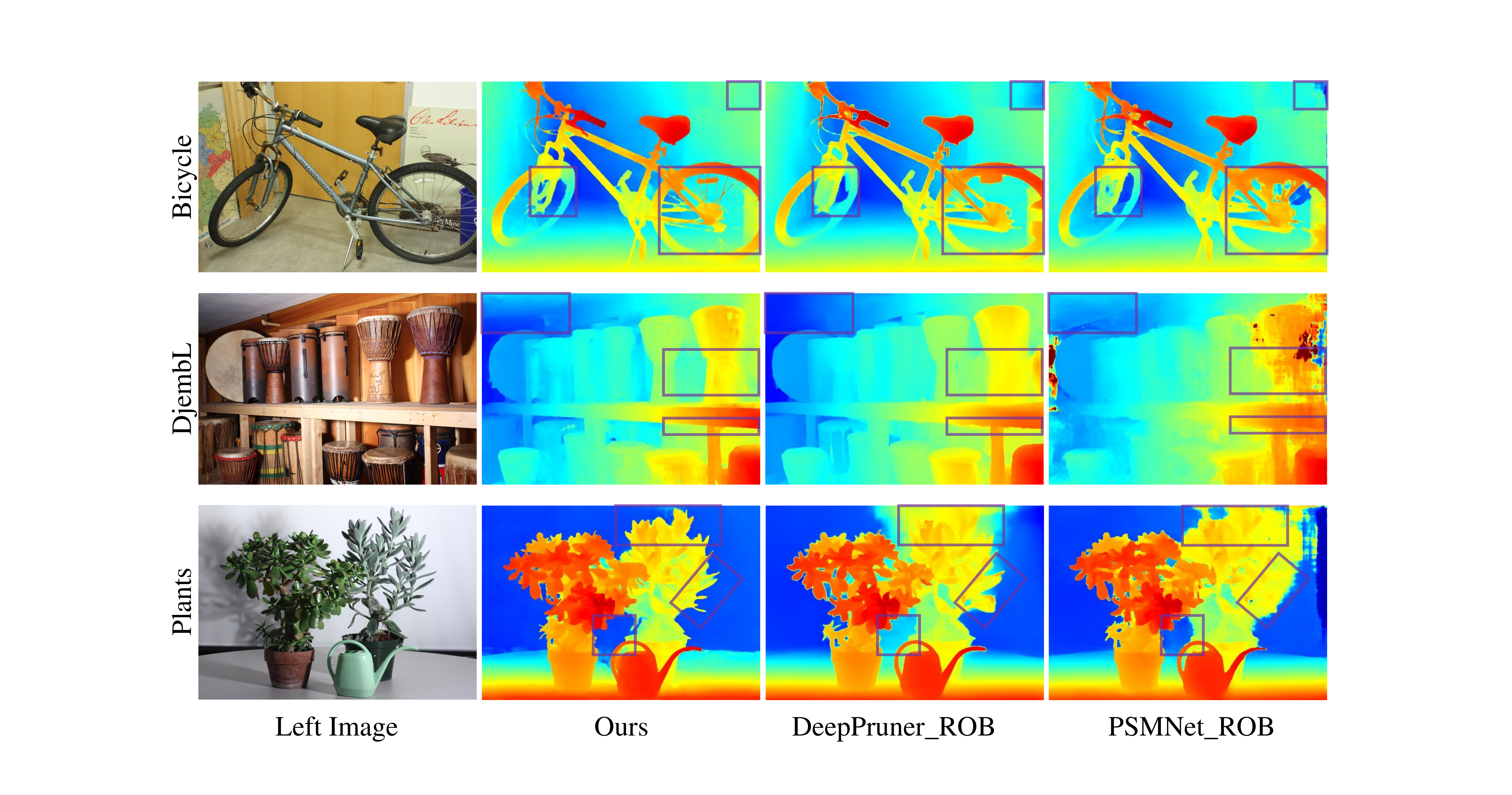}
	\centering
    \caption{The visualization of results on Middlebury-v3. The first column is the left input image. The remaining columns show results for the proposed method, DeepPruner-ROB and PSMNet-ROB respectively.}
    \label{Fig:Middlebury-v3}
\end{figure*}

\subsubsection{Middlebury-v3 Dataset}
Middlebury-v3 \cite{scharstein2014high} is collected in real world with static indoor scenes containing complicated and rich details. Its groundtruth disparities are acquired via structured light, which are both dense and precise. There are 15 stereo pairs for training and 15 stereo pairs for testing. Each pair is provided in three resolutions, full, half and quarter resolution, where we used the quarter resolution in experiment.

In order to better demonstrate the ability of our model in sophisticated and fine-grained indoor scenes, we compare the proposed model with state-of-the-art methods on Middlebury-v3. As we implement our method on the PSMNet and our GPU can only afford the computation cost of quarter resolution when running PSMNet, we mainly focus on the quarter resolution for a fair comparison. As illustrated in Table \ref{tab:Middlebury-v3}, our method achieves the best performance. Compared to PSMNet, our method shows large accuracy improvements but has lower computation cost. For the remaining methods, we achieve better results in most metrics with the same or even the half of their input resolution. Data visualization is also supplied to better present the performance of our method. In Figure \ref{Fig:Middlebury-v3}, there are significant differences between ours and others, especially in the red boxes.



\section{Conclusion}
In this paper, we have presented a content-aware inter-scale cost aggregation method that includes the guidance generation module and the inter-scale cost aggregation module. Our method can not only upsample the cost volume for reliable detail recovery, but also aggregate cross-scale information to improve the quality of disparity estimation. The experiments on Scene Flow, KITTI 2015 and Middlebury-v3 demonstrated that our method improve the performance of stereo matching over many existing state-of-the-art methods.

\section{Acknowledgments}
This work was supported in part by Natural Science Foundation of China (NSFC) under Grants No. 61773062 and No. 61702037.





%
%

\normalem
\bibliographystyle{plain}
\bibliography{MyRefs}
\end{document}